\documentclass[draftclsnofoot,onecolumn]{IEEEtran}
\IEEEoverridecommandlockouts
\usepackage{cite}
\usepackage{amsmath,amssymb,amsfonts}
\usepackage{algorithmic}
\usepackage{hyperref}       
\usepackage{url}            
\usepackage{booktabs} 
\usepackage{graphicx}
\usepackage{textcomp}
\usepackage{multirow}
\usepackage{algorithm}
\usepackage{algorithmic}
\usepackage{array}
\usepackage{bm}
\usepackage{caption3}
\usepackage{url}
\usepackage{pifont} 
\usepackage{subfigure}
\usepackage{wrapfig}
\usepackage{makecell}
\usepackage{colortbl}
\usepackage{enumitem}
\usepackage{xcolor}
\usepackage{authblk}

\def\BibTeX{{\rm B\kern-.05em{\sc i\kern-.025em b}\kern-.08em
    T\kern-.1667em\lower.7ex\hbox{E}\kern-.125emX}}
\begin{document}

\title{Evidential Detection and Tracking Collaboration: New Problem, Benchmark and Algorithm for Robust Anti-UAV System}

\author[1]{\textbf{Xue-Feng Zhu}$^\dag$}
\author[1]{\textbf{Tianyang Xu}$^\dag$$^*$}
\author[2,3,4]{\textbf{Jian Zhao}$^\dag$$^*$}
\author[5]{\textbf{Jia-Wei Liu}$^\dag$}
\author[5]{\textbf{Kai Wang}}
\author[6]{\textbf{Gang Wang}$^*$}
\author[7]{\\\textbf{Jianan Li}$^*$}
\author[8]{\textbf{Qiang Wang}}
\author[9]{\textbf{Lei Jin}}
\author[10]{\textbf{Zheng Zhu}}
\author[11]{\textbf{Junliang Xing}}
\author[1]{\textbf{Xiao-Jun Wu}$^*$}

\affil[1]{Jiangnan University} 
\affil[2]{Institute of North Electronic Equipment}
\affil[3]{Intelligent Game and Decision Laboratory}
\affil[4]{Department of Mathematics and Theories, Peng Cheng Laboratory}
\affil[5]{National University of Singapore}
\affil[6]{Beijing Institute of Basic Medical Sciences}
\affil[7]{Beijing Institute of Technology}
\affil[8]{National Laboratory of Pattern Recognition Institute of Automation, Chinese Academy of Sciences}
\affil[9]{Beijing University of Posts and Telecommunications}
\affil[10]{PhiGent Robotics}
\affil[11]{Tsinghua University}
\maketitle
\renewcommand{\thefootnote}{\fnsymbol{footnote}} 
\footnotetext[1]{Corresponding Authors.}
\footnotetext[2]{Equal Contribution.}
\renewcommand{\thefootnote}{\arabic{footnote}} 
\begin{abstract}

Unmanned Aerial Vehicles (UAVs) have been widely used in many areas, including transportation, surveillance, and military.
However, their potential for safety and privacy violations is an increasing issue and highly limits their broader applications, underscoring the critical importance of UAV perception and defense (anti-UAV). 
Still, previous works have simplified such an anti-UAV task as a tracking problem, where the prior information of UAVs is always provided; such a scheme fails in real-world anti-UAV tasks (\textit{i.e.} complex scenes, indeterminate-appear and -reappear UAVs, and real-time UAV surveillance). 
In this paper, we first formulate a new and practical anti-UAV problem featuring the UAVs perception in complex scenes without prior UAVs information. 
To benchmark such a challenging task, we propose the largest UAV dataset dubbed AntiUAV600 and a new evaluation metric. The AntiUAV600 comprises 600 video sequences of challenging scenes with random, fast, and small-scale UAVs, with over 723K thermal infrared frames densely annotated with bounding boxes. Finally, we develop a novel anti-UAV approach via an evidential collaboration of global UAVs detection and local UAVs tracking, which effectively tackles the proposed problem and can serve as a strong baseline for future research. 
Extensive experiments show our method outperforms SOTA approaches and validate the ability of AntiUAV600 to enhance UAV perception performance due to its large scale and complexity. 
Our dataset, pretrained models, and source codes will be released publically.

\end{abstract}

\section{Introduction}
\label{introduction}

Due to the significant advantages of high autonomy and low cost, Unmanned Aerial Vehicles (UAVs) have been widely utilized across numerous fields~\cite{vskrinjar2019application, del2021unmanned, partheepan2023autonomous}. Despite promising, UAVs raise potential risks for violating public safety and personal privacy, highly impacting their broader applications. Therefore, the UAV perception and defense (anti-UAV) is particularly important for reliable open-world UAV applications and has received emerging research interests within the computer vision community \cite{zhao2022vision, zhao20233rd}. However, current works have simplified such an anti-UAV task as a tracking problem, where the prior information of UAVs is always provided; such a scheme fails in open-world anti-UAV tasks where UAVs can randomly appear in complex scenes at any time.

\begin{figure}[t]
\begin{center}
\includegraphics[trim={18mm 81mm 20mm 0mm},clip,width=0.95\linewidth]{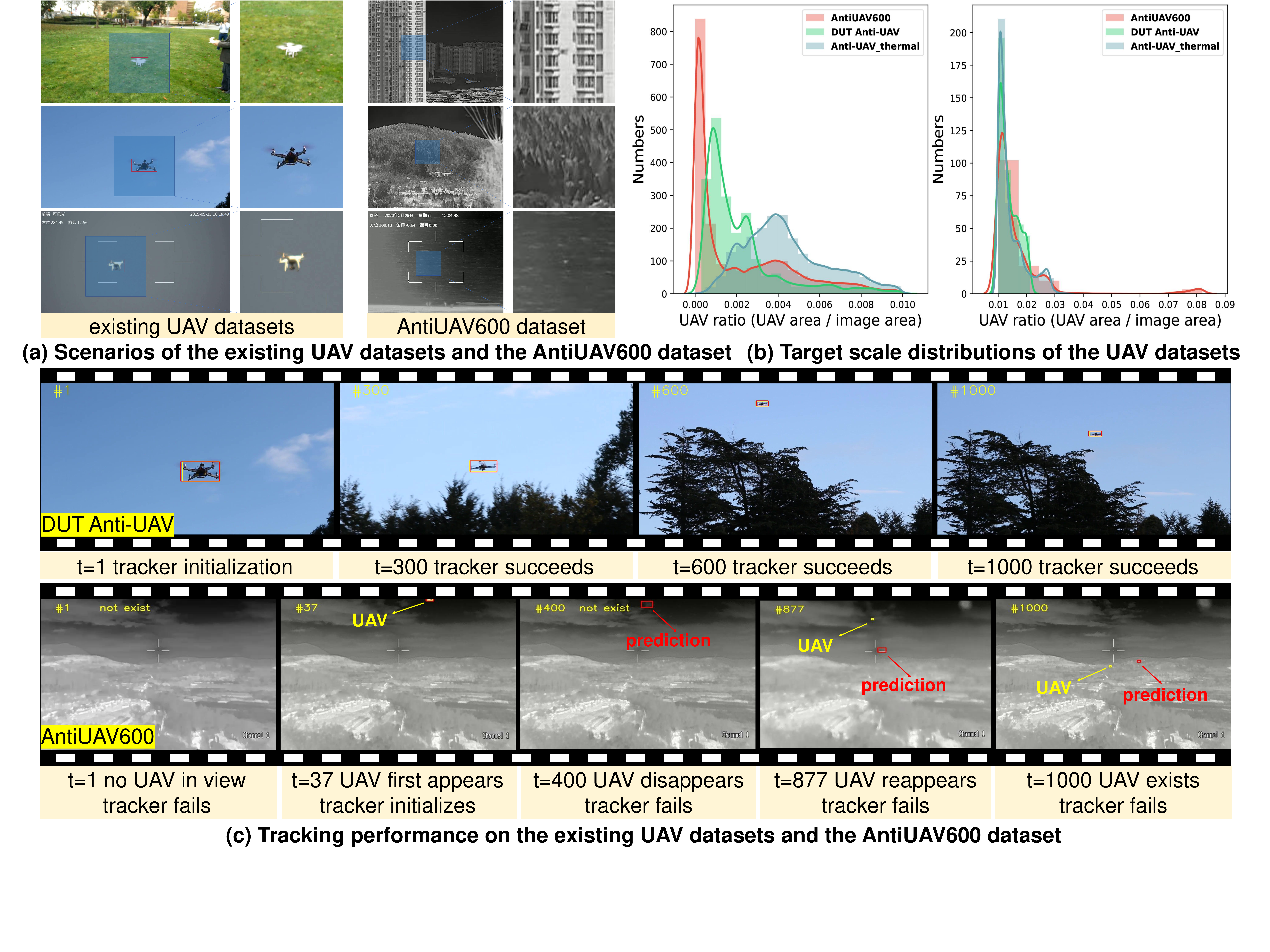}
\end{center}
\caption{Comparison of the proposed AntiUAV600 dataset and the existing UAV datasets. The proposed AntiUAV600 dataset offers several advantages over existing UAV datasets. Firstly, it features more complex scenarios, as depicted in subfigure (a). Secondly, the UAVs in AntiUAV600 have a smaller average scale and a larger scale variation, as shown in subfigure (b). Finally, existing tracking methods struggle to effectively handle the challenges presented in the proposed AntiUAV dataset, as demonstrated in subfigure (c).}\label{motivation}
\end{figure}

To support practical open-world anti-UAV perception, we for the first time propose a new anti-UAV task featuring UAVs perception in complex scenes without any prior UAV information. This task necessitates continuously detecting and tracking UAV targets in real-time complex video streams without relying on the first-frame UAV template which in contrast is a prerequisite for previous works~\cite{jiang2021anti, zhao2022vision, huang2021siamsta}. However, existing anti-UAV datasets~\cite{jiang2021anti,zhao2022vision} cannot favor such a practical task due to three reasons. Firstly, they require the UAV target to appear in the first frame of the video to serve as the input template for UAV tracking, which fails to reflect the challenging real-world scenarios where UAVs can appear anywhere and at any time. Secondly, the UAVs of these datasets~\cite{zhao2022vision, coluccia2021drone} occupy large regions of images, making it infeasible for real-world UAVs surveillance when the UAVs occupy very few pixels, as shown in Fig.~\ref{motivation}(b). Thirdly, both the quality and scale of existing datasets are limited for evaluating real-world anti-UAV tasks.


To tackle these limitations and benchmark the proposed anti-UAV task, we propose a new large-scale anti-UAV dataset, dubbed AntiUAV600, and a novel evaluation metric. To the best of our knowledge, the AntiUAV600 is currently the largest UAV dataset in computer vision. It contains 600 thermal infrared video sequences with over 720K densely annotated frames. These sequences are captured in complex outdoor scenes with new UAV challenges such as random UAVs appearing, fast motion, thermal crossovers, and tiny UAV scales. As exampled in Fig.~\ref{motivation}(c), the AntiUAV600 contains many practical real-world scenarios where the UAV targets do not appear at the beginning of videos, which existing tracking approaches fail to tackle with. Accordingly, a new evaluation metric combining the average Intersection over Union (IoU) score with a penalty item is developed, to evaluate the anti-UAV systems' ability to localize UAV targets and predicting their absence in video streams.


The AntiUAV600 opens practical and challenging research directions that can not be effectively tackled by existing UAV tracking-based approaches~\cite{zhao20212nd, jiang2021anti, zhao2022vision}. To this end, we propose a novel framework for continuously detecting and tracking UAV targets. In order to tackle the random appearance and absence problem of UAVs in real-world videos, we propose a novel evidential collaboration framework of the global UAVs detections and local UAVs tracking. The framework starts from detecting UAV targets in real-world videos and switch to the tracking branch once any UAV target is localized. Then, we develop a new tracking backbone with a novel additional evidential head to predict the confidence of tracked UAVs, based on which our framework can smoothly switch from the tracking branch to the detection branch. With such adaptive switching of detection and tracking branches, our proposed framework achieves SOTA performance on the challenging AntiUAV600 and serves as a strong baseline for future research. 


Our main contributions are summarised as follows:
\vspace{-1mm}
\begin{itemize}[leftmargin=0cm, itemindent=0.3cm, itemsep=1.0mm]
\item We propose a new real-world anti-UAV problem and a challenging benchmark with the largest UAV dataset and a new evaluation measure.
\item We propose a novel anti-UAV framework with the evidential collaboration of global UAVs detection and local UAVs tracking.
\item Extensive experiments and analyses demonstrate the superiority of the proposed AntiUAV600 dataset and the framework.
\end{itemize}

\section{Related Works}
\label{related_works}                                       
\textbf{Anti-UAV Datasets and Methodologies.} 
In recent years, deep learning techniques have been applied to address the challenge of anti-UAV task~\cite{isaac2021unmanned, dadboud2021single, xie2023stftrack}, leading to the development of various UAV datasets for training and evaluation purposes.
The Drone-vs-Bird Detection Challenge at IEEE AVSS2021 \cite{coluccia2021drone} provides a UAV detection dataset comprised of 77 videos for training and 21 videos for evaluation.
In this challenge, the best methods are YOLOv5-based detectors, with additional modeling tricks to boost the performance.
Besides, the DUT Anti-UAV~\cite{zhao2022vision} dataset presents 10K visible images for UAV detection and 20 video sequences for UAV tracking.
A simple UAV tracking method is proposed that utilizes global detection to re-detect a lost target.
Additionally, a large-scale dataset named Anti-UAV~\cite{jiang2021anti} provides 318 videos with over 297K annotated RGB-Thermal frames for UAV tracking.
Simultaneously, a novel UAV tracking approach is proposed, which utilizes dual-flow semantic consistency strategies for discriminative feature learning.
In addition, based on the Anti-UAV dataset, Huang~\textit{et al.}~\cite{huang2021siamsta} propose a siamese-based tracker with introduced spatio-temporal attention and re-detection mechanism for tracking UAVs.
In the recently held Anti-UAV Challenges~\cite{zhao20212nd, zhao20233rd}, advanced methods for multi-scale UAV tracking have been developed.
Li~\textit{et al.}~\cite{li2023global} propose a global-local UAV tracking method that integrates both appearance and motion models.
Yu~\textit{et al.}~\cite{yu2023unified} devise a robust UAV tracker by combining multi-region local tracking and global detection, incorporating background correlation and dynamic small target detection to address the challenges of dynamic background and small UAV scales.
Despite numerous proposed datasets and algorithms, the existing anti-UAV datasets and solutions are typically served separately for either detection or tracking.
While in practical scenarios, the anti-UAV system requires adaptive detection and continuous tracking of the UAV targets. 
Therefore, this work formulates a new anti-UAV task, a large-scale benchmark, and a novel solution.
 
\textbf{Evidential Deep Learning (EDL).} EDL is a framework for incorporating uncertainty and reasoning under uncertainty into deep learning models~\cite{shi2020multifaceted, sensoy2020uncertainty}. 
It allows for the modeling of uncertain data and the propagation of this uncertainty throughout the learning process, providing a way to quantify the uncertainty of the model's predictions.
Recently, the evidential theory has been increasingly incorporated into deep neural network learning for classification~\cite{sensoy2018evidential, bao2021evidential} and regression~\cite{amini2020deep} tasks, providing improved robustness via quality assessment.
EDL enables the model to distinguish between known and unknown data. 
The learned predictive uncertainty can be utilized as a confidence scoring function to identify out-of-distribution samples.
For anti-drone systems, determining whether a local tracker effectively tracks a drone target and switching to global detection is crucial.
Therefore, in our work, we design an evidential head as a switcher to identify the uncertainty of the tracking results and enable adaptive collaboration between detection and tracking.

\section{New Anti-UAV Benchmark}
\label{dataset}
Real-world UAV perception is particularly challenging due to the UAVs randomly appearing or re-appearing from the system's field of view at any time and any location. 
Recently, significant progress~\cite{zhao20212nd, zhao2022vision} has been witnessed by approaching the anti-UAV task as a UAV tracking problem.
The UAV tracking paradigm can be defined as $\bm{x}_t^\ast=\sideset{}{}{\arg\max}_{\bm{x}_t}{f(\bm{X}_t; \bm{Z}, \bm{w})}$, where $\bm{X}_t$ is the observation of current $t$-th frame, $\bm{Z}$ is the given UAV template, and $f(.)$ is the network with weights $\bm{w}$.
The paradigm aims to determine the optimal target candidate $\bm{x}_t^\ast$ by predicting the observation based on the given target state and learned network.
However, acquiring the UAV template $\bm{Z}$ as prior information in advance is impractical for real-world anti-UAV systems, posing a challenge for the application of UAV tracking in real-time UAV surveillance.

To advance the research of anti-UAV task and support their real-world applications, we formulate a new anti-UAV problem: continuously detecting and tracking UAV targets in real-time complex video streams without relying on the first-frame UAV template. Given real-time captured videos, the anti-UAV systems are supposed to perceive the presence or absence of UAVs and accurately localize them in case of their presence. The task can be formulated as $\bm{x}_t^\ast=\sideset{}{}{\arg\max}_{\bm{x}_t}{f(\bm{X}_t; \bm{w})}$, which is more challenging and applicable to a wider range of real-world scenarios.
To support and benchmark the proposed new anti-UAV problem, we introduce a new high-quality anti-UAV benchmark composed of a UAV-targeted dataset AntiUAV600 and a new evaluation metric.

\subsection{Dataset Details}
The AntiUAV600 dataset comprises $300$ training sequences, $50$ validation sequences, and $250$ test sequences, with $337$K thermal infrared frames in the training set, $56$K in the validation set and $330$K in the test set. 
All frames were captured using a thermal infrared camera in a resolution of $640\times512$ at a frame rate of $25$ frames per second (fps).

Fig.~\ref{motivation}(a-c) illustrates the comparison of our AntiUAV600 and existing datasets, where our dataset is much more challenging in (1) complex backgrounds, (2) small UAV scales, and (3) random UAV appearance / re-appearance in videos. AntiUAV600 covers various challenging background scenes including the sky with clouds, city buildings, mountains, rivers, and forests, etc, supporting more practical real-world applications. In addition, as exampled in Fig.~\ref{motivation}(c), previous UAV tracking approaches require the UAV to always appear in the first frame as a prior template, which is infeasible for practical real-world applications. In contrast, our AntiUAV600 features UAV targets random presence and absence throughout the video, posing new challenges for anti-UAV approaches.
Please refer to the supplementary material for more details of the proposed AntiUAV600 dataset.

\begin{figure}[t]
\centering
\subfigure[Train]{
\begin{minipage}[t]{0.24\linewidth}
\centering
\includegraphics[trim={23mm 12mm 50mm 28mm},clip,width=1\linewidth]{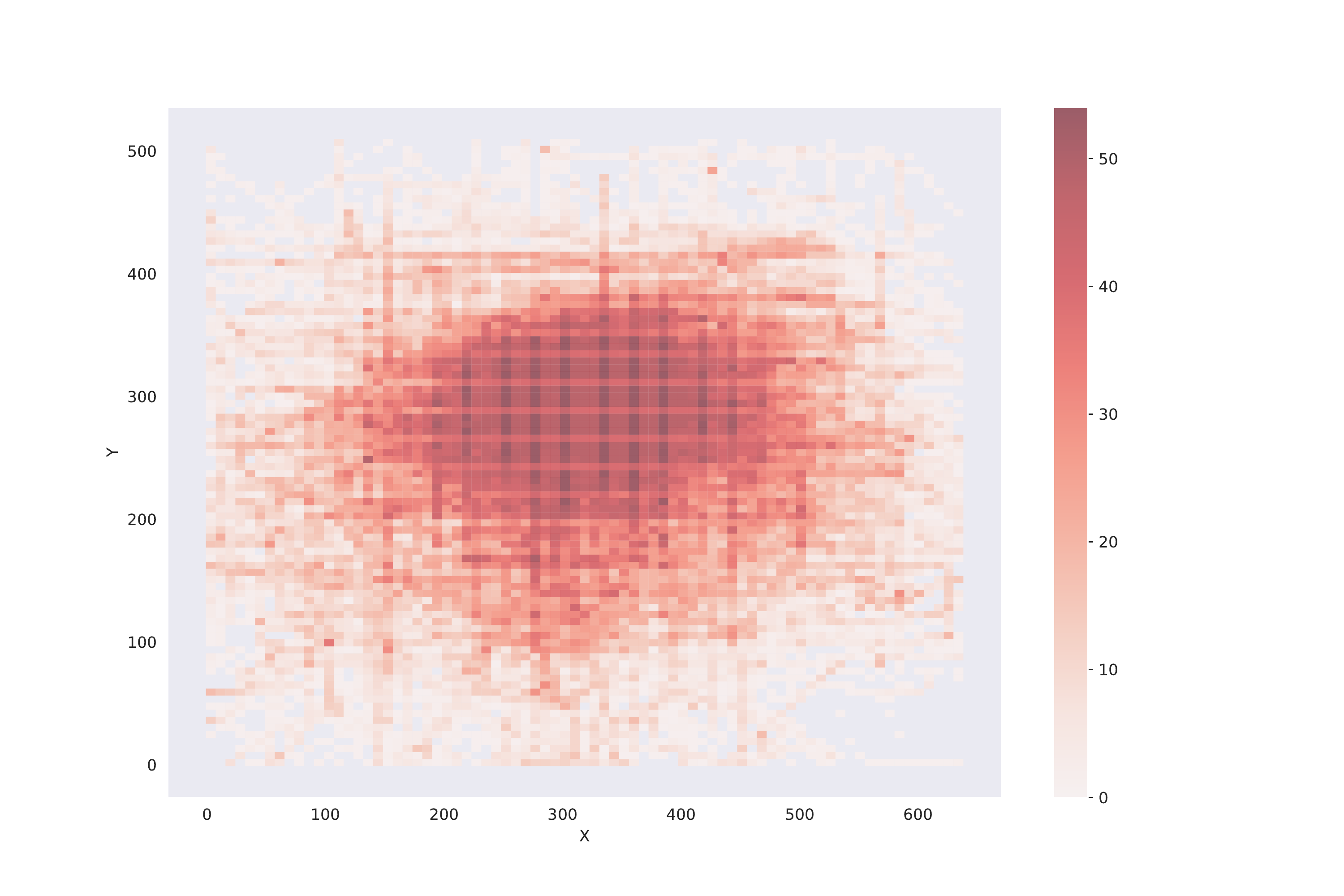}
\end{minipage}%
}%
\subfigure[Val]{
\begin{minipage}[t]{0.24\linewidth}
\centering
\includegraphics[trim={23mm 12mm 50mm 28mm},clip,width=1\linewidth]{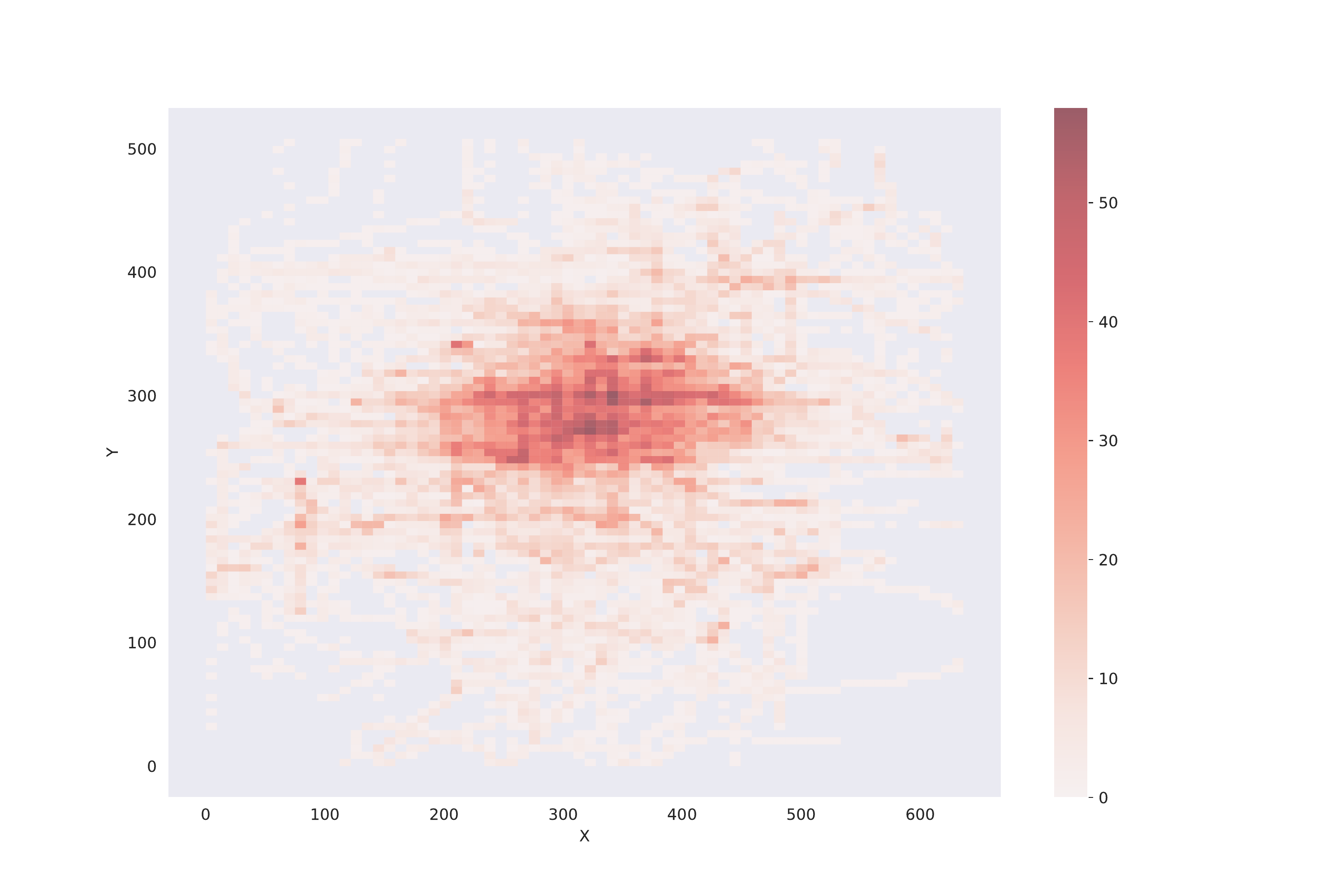}
\end{minipage}%
}%
\subfigure[Test]{
\begin{minipage}[t]{0.24\linewidth}
\centering
\includegraphics[trim={23mm 12mm 50mm 28mm},clip,width=1\linewidth]{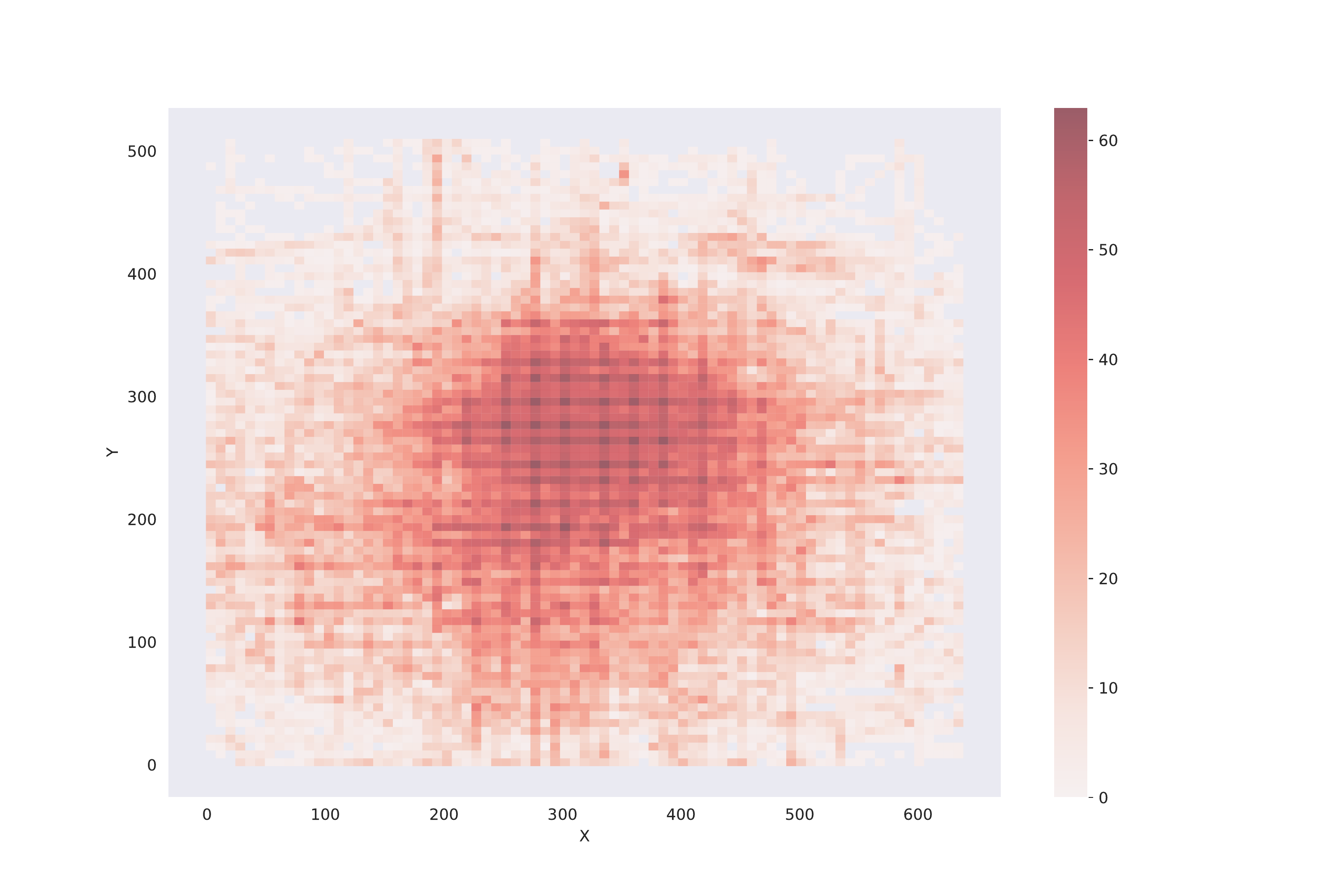}
\end{minipage}
}%
\subfigure[Scale]{
\begin{minipage}[t]{0.24\linewidth}
\centering
\includegraphics[trim={10mm 5mm 10mm 10mm},clip,width=1\linewidth]{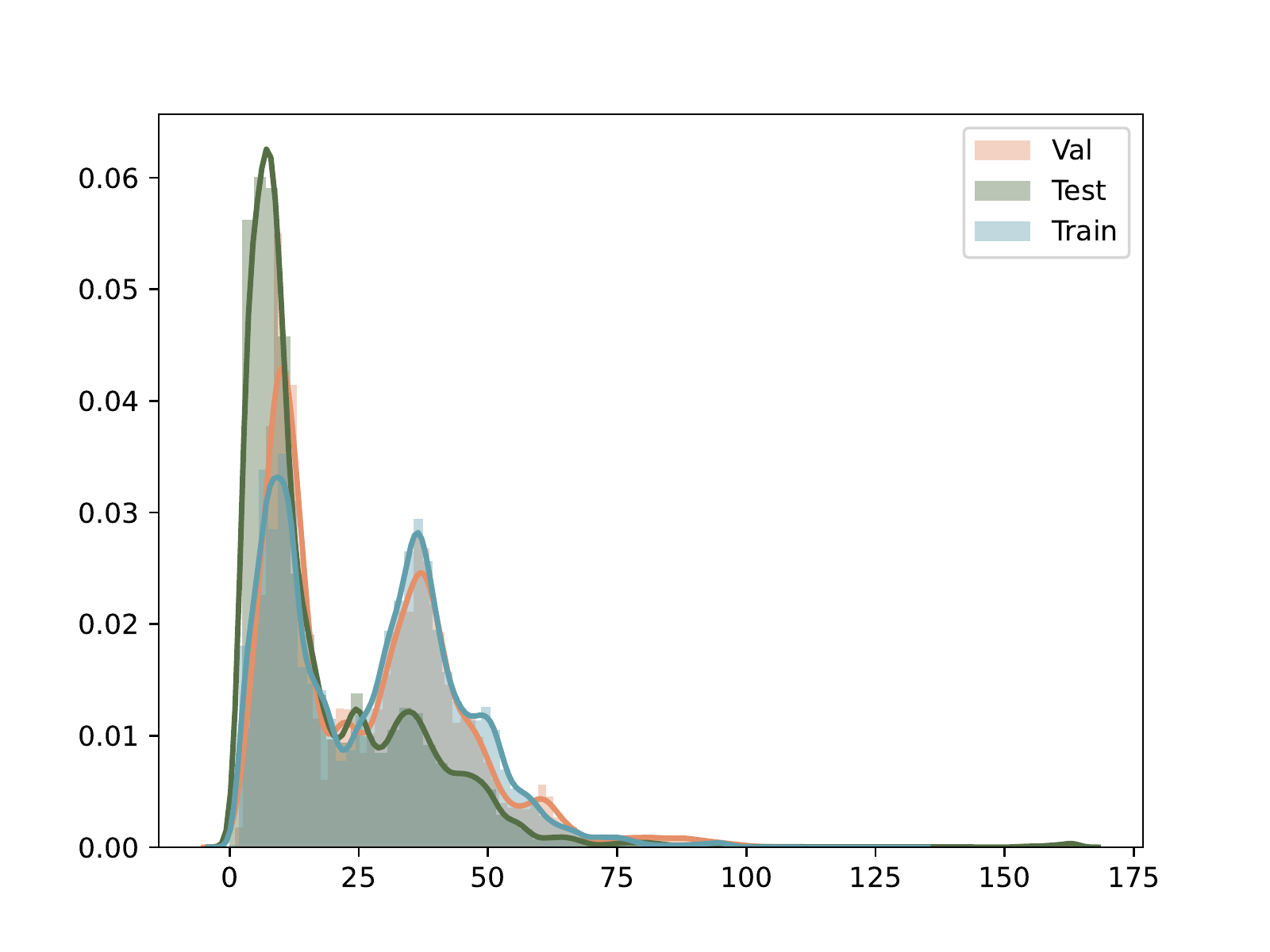}
\end{minipage}
}%
\centering
\caption{The position and scale distribution of the AntiUAV600 dataset.}
\label{position_scale}
\end{figure}

\begin{table}[t]
\footnotesize
\centering
\caption{A comparison of AntiUAV600 benchmark with related UAV detection and tracking datasets.}\label{table1}
\resizebox{0.95\linewidth}{!}{
\begin{tabular}{cccccc}
\hline \textbf{Dataset} & \textbf{\#Sequences} & \textbf{\#Frames} & \textbf{\#Avg. UAV Size ($\sqrt{wh}$)} &\textbf{\#Attributes} & \textbf{Modality} \\
\hline 
USC Drone~\cite{chen2017deep} & 60 & 20K & 114$\times$83 (97)& - & RGB \\
Halmstad Drone~\cite{svanstrom2021real} & 650 & 203K& - & - & RGB, Thermal \\
MAV-VID~\cite{rodriguez2020adaptive} & 64 & 40K& 215$\times$128 (166) & - & RGB\\
Drone-vs-Bird Challenge~\cite{coluccia2021drone} & 77 & 105K& 34$\times$23 (28) & - & RGB\\
Anti-UAV~\cite{jiang2021anti} & 318 & 297K& \makecell[c]{RGB:125$\times$63 (89)\\IR:52$\times$30 (39)} & 7 & RGB-Thermal\\
DUT Anti-UAV~\cite{zhao2022vision} & 20 & 35K& 70$\times$31 (47) & - & RGB\\
\rowcolor{gray!20}AntiUAV600 & 600 & 723K & 30×19 (23) & 7 & Thermal\\
\hline
\end{tabular}}
\end{table}

\subsection{Data Annotation}
The AntiUAV600 is densely annotated with UAVs targets bounding boxes for frames containing UAVs, and frames without UAVs are labeled with 'Not Exist'. In addition, all frames are annotated with seven challenging attributes, including Out-of-View (OV), Occlusion (OC), Fast Motion (FM), Scale Variation (SV), Infrared Crossover (IC), Dynamic Background Clusters (DBC) and Target Scale (TS). 
The detailed definitions and distributions of the attributes are presented in the supplementary material.

As shown in Tab.~\ref{table1}, our AntiUAV600 is the largest UAV dataset and has the smallest average target size compared to other UAV datasets (Fig.~\ref{motivation}(b)), making it much more challenging for UAV perception.
Besides, we illustrate the statistics for the distribution of UAVs positions and scales of AntiUAV600 in Fig.~\ref{position_scale}, where (a-c) illustrate the position distribution of UAVs in the training, validation, and test set, respectively.
As depicted, the movement of UAVs in our dataset is diverse, with their traces scattered throughout the entire field of view.
Fig.~\ref{position_scale} (d) shows the scale distributions of UAVs in the three subsets, which is calculated by $\sqrt{w\times{h}}$ from the width $w$ and height $h$ of the bounding box.
The training, validation, and test sets share a similar scale distribution with a very small UAVs average scale, which poses new significant challenges for the anti-UAV tasks.

\subsection{Evaluation Metric}
To evaluate the challenging anti-UAV problem, we propose a novel performance metric named Acc,
\begin{small}
\begin{equation}\label{Acc}
Acc = \sum_{t=1}^{T}\frac{{\mathrm{IoU}}_t\times\delta(v_t>0)+p_t\times(1-\delta(v_t>0))}{T}-\alpha\times(\sum_{t=1}^{T^*}\frac{q_t\times\delta(v_t>0)}{T^*})^\beta,
\end{equation}
\end{small}
where $\mathrm{IoU}_t$ is the IoU between the predicted and the ground-truth boxes at the $t$-th frame.
$p_t$ denotes the predicted visibility flag ($1$ for empty predicted box, $0$ otherwise), $q_t$ represents the prediction failure flag ($1$ for localization failure, $0$ otherwise), and $v_t$ is the ground-truth visibility flag ($1$ when $v_t>0$, $0$ otherwise).
The accuracy is averaged over all frames in a sequence; $T$ denotes the number of total frames and $T^*$ denotes the number of frames where the UAV target is visible.

The first term of the proposed metric measures the accuracy between the predicted boxes and ground-truth boxes when the target is visible, using the Intersection over Union (IoU) metric. It can also evaluate the methods' performances of detecting the absence of a UAV target.
The second term is a penalty that penalizes cases where the UAV target exists but the prediction fails to catch any target region.
This term is crucial for evaluating anti-UAV systems because failures to detect UAV targets' presence can result in high costs in practical applications.

\section{Evidential Detection and Tracking Collaboration for Anti-UAV Solution}
\label{approach}
To inspire new designs for anti-UAV systems and to establish a baseline method for comparisons of future works, we introduce a new anti-UAV algorithm for our proposed real-world anti-UAV task, \emph{i.e.}, Evidential Detection and Tracking Collaboration (EDTC). As shown in Fig.~\ref{pipeline}, our EDTC comprises of a global detection branch to detect UAV targets across the entire field of view, and a local tracking branch to continuously localize the targets in video streams. These two branches are adaptively switched to achieve efficient and accurate UAV perception for real-world complex videos without relying on prior UAVs information.

\subsection{Global Detection} 
\begin{figure}[t]
\begin{center}
\includegraphics[trim={0mm 90mm 00mm 0mm},clip,width=0.95\linewidth]{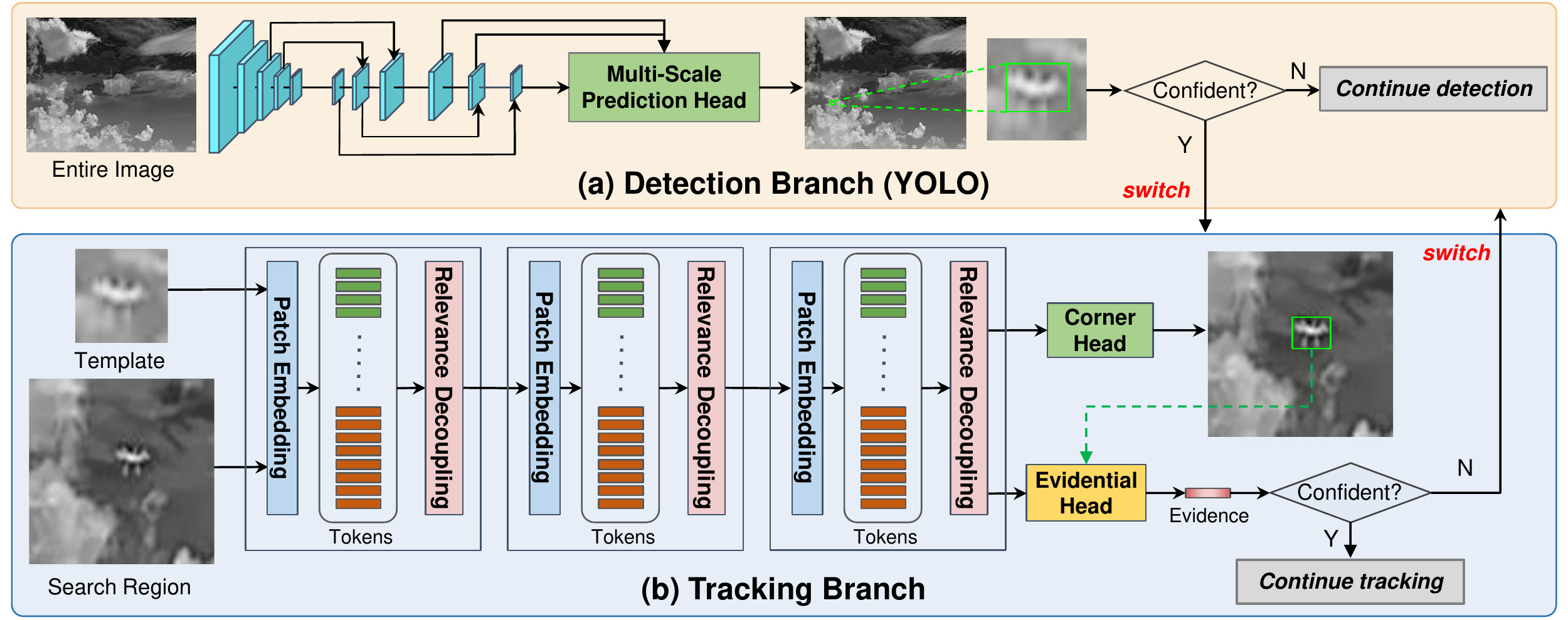}
\end{center}
\caption{The pipeline of the proposed anti-UAV approach. The adaptive switching between the global detection and local tracking branches delivers a robust anti-UAV visual system.}\label{pipeline}
\end{figure}

Since our new anti-UAV problem features continuously detecting and localising UAVs without relying on prior first-frame UAV presence, we propose to first use a UAV-specific detector, \emph{i.e.}, YOLOv5s model in Fig.~\ref{pipeline}(a), to detect global UAVs throughout the videos. Once the UAV target is detected, our EDTC can smoothly switch to the local tracking branch to continuously track the detected target UAV until its absence. If the local tracker returns a denied tracking result, our EDTC will automatically activate the global detector again to perceive the UAVs re-appearance throughout the entire video frames and repeat the above procedures until the end of the video.




\begin{figure}[t]
\begin{center}
\includegraphics[trim={0mm 113mm 00mm 0mm},clip,width=0.95\linewidth]{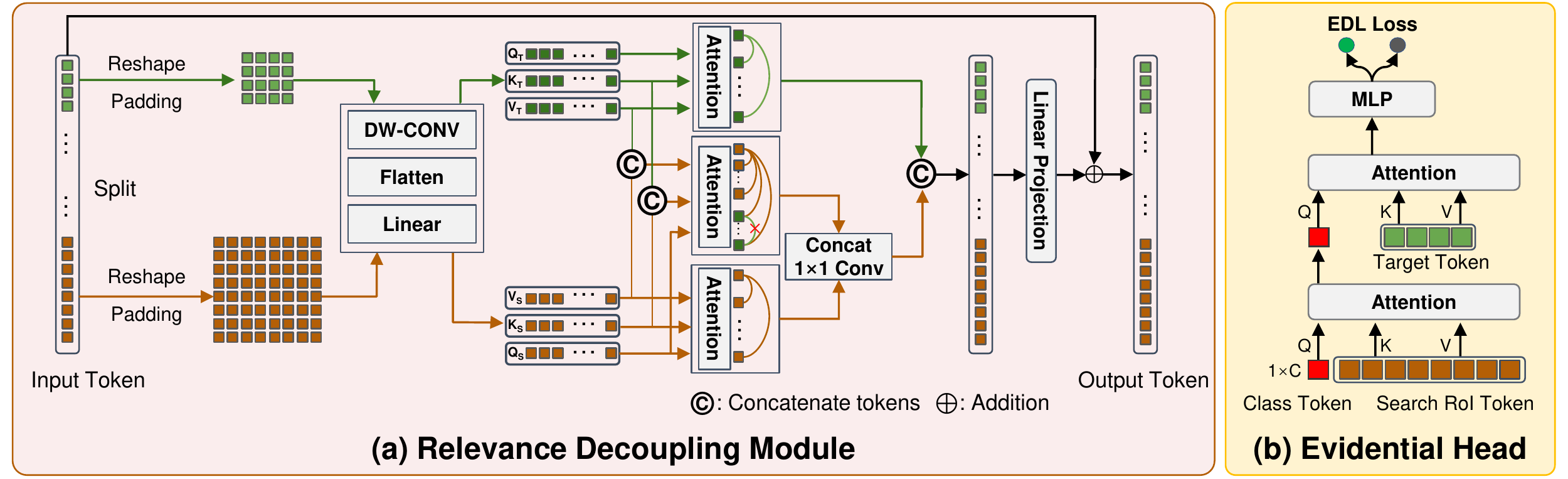}
\end{center}
\caption{The proposed relevance decoupling module and evidential head in the tracking branch.}\label{RDM}
\end{figure}

\subsection{Local Tracking} 
\textbf{Tracking Framework.} We propose a novel deep evidential tracking approach that comprises a backbone network stacking a three-stage relevance decoupling module, a corner head predicting the target bounding box, and an evidential head that learns the predictive uncertainty.
As shown in Fig.~\ref{pipeline}, the multi-stage backbone network extracts and integrates the features of the template and search region. 
For each stage of backbone, a patch embedding composed of a convolutional layer is utilized to map the image patches to feature tokens, which can reduce the spatial resolution and aggregate the local patterns into the channel dimension, implicitly modelling the 2D spatial smoothness.
Then, the tokens of templates and search region are flattened and concatenated and fed into the relevance decoupling module for feature extraction and matching.
Finally, the search region features are fed into the corner prediction head~\cite{yan2021learning} to estimate the target position and scale.
The training loss is a combination of IoU loss and $L_1$ loss, as $L=\lambda_{iou}L_{iou}(\bm{b}_i, \hat{\bm{b}}_i) + \lambda_{L_1}L_{1}(\bm{b}_i, \hat{\bm{b}}_i)$,
\noindent where $\bm{b}_i$ and $\hat{\bm{b}}_i$ are the ground-truth and predicted boxes, and $\lambda_{iou}$ and $\lambda_{L_1}$ are the loss weights.
For more details of the tracking branch, please refer to the supplementary material.

\textbf{Relevance Decoupling Module.} The relevance decoupling module aims to extract and integrate the features from templates and search instances. 
As illustrated in Fig.~\ref{RDM}, the concatenated tokens of templates and search regions are split into two parts and reshaped to feature maps to allow for depth-wise convolutional operation to extract local spatial information.
Then the feature maps are flattened and processed by a linear projection to generate query, key, and value tokens for attention operations.
Specifically, the template tokens are represented as $\bm{Q}_t$, $\bm{K}_t$, and $\bm{V}_t$, while the search tokens are represented as $\bm{Q}_s$, $\bm{K}_s$, and $\bm{V}_s$.
The attention operations are calculated as follows:
\begin{small}
\begin{equation}\label{attention}
\begin{aligned}
&\mathrm{SelfAttn}_t= \mathrm{Softmax}(\frac{\bm{Q}_t\bm{K}_t^T}{\sqrt{d}})\bm{V}_t,
&\mathrm{SelfAttn}_s= \mathrm{Softmax}(\frac{\bm{Q}_s\bm{K}_s^T}{\sqrt{d}})\bm{V}_s,\\
&\mathrm{CrossAttn}_{ts}= \mathrm{Softmax}(\frac{\bm{Q}_s\bm{K}_m^T}{\sqrt{d}})\bm{V}_m,
&\bm{K}_m = \mathrm{Concat}(\bm{K}_t, \bm{K}_s), \bm{V}_m = \mathrm{Concat}(\bm{V}_t, \bm{V}_s),
\end{aligned}
\end{equation}
\end{small}
\noindent where the $\mathrm{SelfAttn_t}$ and $\mathrm{SelfAttn_s}$ are the self-attention operations of the template and search tokens, respectively.
The $\mathrm{CrossAttn_{ts}}$ is the cross-attention operation for the template and search integration.
The output of $\mathrm{SelfAttn_s}$ contains only the search region-independent features, which can be used for UAV target-specific local detection. 
On the other hand, the outputs of $\mathrm{CrossAttn_{ts}}$ are the matched template and search features, which are used for template matching-based local tracking.
Therefore, the proposed tracking branch can achieve local detection and tracking of UAVs, resulting in more stable UAV-awareness results.

Subsequently, the search region and integrated features are concatenated across the channel dimension, following a $1\times1$ convolutional layer to reduce the dimension.
Finally, the template and search tokens are concatenated and passed through a linear projection followed by an addition operation to fuse the input tokens with the extracted features.

\textbf{Evidential Head.} 
The most fundamental goal of an anti-UAV system is to reduce tracking failures that may arise from various challenging scenarios.
Therefore, predicting the tracking uncertainty is crucial for anti-UAV systems, as the uncertainty can be used as a scoring function for judging tracking results.
Hence, we propose an evidential head to evaluate the evidence of the prediction, as depicted in Fig.~\ref{RDM}.
Inspired by the Ref.~\cite{sensoy2018evidential} for image classification and uncertainty modeling, we assume the class probability $\bm{p}_i$ of sample $\bm{x}_i$ for $K$-class classification follows a Dirichlet distribution, and the loss function for evidence $\bm{e}_i$ learning is,
\begin{small}
\begin{equation}\label{evidential loss}
\mathcal{L}_i(\bm{w}) = \sum\nolimits_{k=1}^{K}\bm{y}_{ik}(\log(S_i)-\log(\bm{\alpha}_{ik})),
\end{equation}
\end{small}
\noindent where the $\bm{y}_{i}$ is the one-hot vector of the ground-truth for the sample $\bm{x}_i$.
The $\bm{e}_i$ can be represented as $\mathrm{RELU}(f(\bm{x}_i|\bm{w}))$ and $f(.)$ is the network with weights $\bm{w}$.
The RELU activation guarantees a non-negative value for the output evidence.
Besides, the Dirichelet distribution parameters $\bm{\alpha}_i$ are related to the evidence as $\bm{\alpha}_i = \bm{e}_i + 1$ and the Dirichelet strength is defined as $S_i = \sum_{k=1}^{K}\bm{e}_{ik}$.
In the inference stage, the probability of the $k$-th class $\hat{\bm{p}}_k$ can be predicted as $\hat{\bm{p}}_k=\bm{\alpha}_k/S$ and the predictive uncertainty $u$ can be calculated as $u = K/S$.

In our framework, we evaluate the tracking result as a binary classification problem to determine whether the tracking results belong to the target or background. As shown in Fig.~\ref{RDM}(b), the proposed evidential head is composed of two attention modules and a multi-layer perceptron.
Firstly, we define a learnable class token as a query for the attention of tokens of search ROI region cropped using the predicted target bounding box.
The class token allows the encoding of the tracking prediction information.
Then, the class token is input as a query for the attention with target template tokens to compare the encoded prediction information with the initial target.
Finally, the evidence vector is predicted by an MLP layer and a RELU layer.
With the learned evidence, the class probability and uncertainty of the tracking prediction can be determined.
If the class probability indicates a target class with uncertainty lower than $\theta_{eh}$, the system continues tracking in the next frame; otherwise, the system switches to global detection.


\section{Experiments}
\label{evaluation}

\subsection{Implementation Details}
The detection branch and the tracking branch are trained separately.
The YOLOv5s model is trained on the AntiUAV600 training set, while the tracking model is trained on our AntiUAV600 dataset and four general tracking datasets including LaSOT~\cite{fan2019lasot}, TrackingNet~\cite{muller2018trackingnet}, GOT10K~\cite{huang2019got} and COCO~\cite{lin2014microsoft}.
The inference speed of the proposed EDTC is about $35$ fps.
The $\alpha$ and $\beta$ in Eq.~\ref{Acc} are set to $0.2$ and $0.3$, respectively. 
More implementation details and settings are provided in the supplementary material.


\subsection{Comparative Experiments}
\textbf{Performance on AntiUAV600.} We report the quantitative comparisons of our method and various constructed detection-tracking anti-UAV methods in Tab.~\ref{comparative}. The compared methods are built via a sequential combination of detection and SOTA tracking methods. 
Specifically, a UAV-specific detector first detects the initial UAV target, then a tracking method tracks the detected UAV.
We select transformer-based, Siamese-based and filter-based trackers, including OSTrack~\cite{ye2022joint}, MixFormer~\cite{cui2022mixformer}, ToMP~\cite{mayer2022transforming}, Stark~\cite{yan2021learning}, KeepTrack~\cite{mayer2021learning}, SiamCAR~\cite{guo2020siamcar} and DiMP~\cite{bhat2019learning}, and combine them with the trained UAV-specific YOLOv5s model.
The EDTC and EDTC$^\star$ respectively represent the tracking branch of the proposed method without the use of UAV data and with the use of UAV data for training.
As presented in Tab.~\ref{comparative}, our proposed evidential collaboration of detection and tracking achieves significantly superior performance compared to other approaches.



\begin{table}[!t]
\footnotesize
\centering
\caption{Results of the comparative experiments on AntiUAV600.}\label{comparative}
\resizebox{0.95\linewidth}{!}{
\begin{tabular}{cccccccccc}
\hline Method (YOLO + ) & OSTrack & MixFormer & ToMP &  Stark & KeepTrack & SiamCAR & DiMP & EDTC & EDTC$^\star$ \\
\hline 
Acc on AntiUAV600   & 0.280 & 0.261 & 0.196 & 0.264 & 0.329 & 0.194 & 0.144 & 0.439 & 0.486\\
\hline
\end{tabular}}
\end{table}

\begin{table}[!t]
\footnotesize
\centering
\caption{Results of the comparative experiments on DUT Anti-UAV and Anti-UAV datasets.}\label{comparative2}
\resizebox{0.95\linewidth}{!}{
\begin{tabular}{cccccccccc}
\hline Method & OSTrack & MixFormer & ToMP & Stark & KeepTrack & SiamCAR & DiMP & EDTC & EDTC$^\star$ \\
\hline 
Initial bounding box   & \ding{52} & \ding{52} & \ding{52} & \ding{52} & \ding{52} & \ding{52} & \ding{52} & \ding{56} & \ding{56}\\
Acc on DUT Anti-UAV  & 0.564 & 0.546 & 0.566 & 0.531 & 0.592 & 0.461 & 0.560 & 0.593 & 0.606\\
Acc on Anti-UAV      & 0.549 & 0.503 & 0.494 & 0.521 & 0.519 & 0.389 & 0.418 & 0.617 & 0.634 \\
\hline
\end{tabular}}
\end{table}

\begin{table}[!t]
\footnotesize
\centering
\caption{Results on the scenarios with UAV targets experiencing absence and reappearance.}\label{results_asence}
\resizebox{0.95\linewidth}{!}{
\begin{tabular}{cccccccccc}
\hline Method & OSTrack & MixFormer & ToMP & Stark & KeepTrack & SiamCAR & DiMP & EDTC & EDTC$^\star$ \\
\hline 
Initial bounding box   & \ding{52} & \ding{52} & \ding{52} & \ding{52} & \ding{52} & \ding{52} & \ding{52} & \ding{56} & \ding{56}\\
Acc  & 0.035 & 0 & 0 & 0.001 & 0.074 & 0 & 0.008 & 0.371 & 0.387\\
\hline
\end{tabular}}
\end{table}

\textbf{Performance on other UAV Datasets.} To further demonstrate the effectiveness of the proposed baseline method EDTC for UAV perception, we evaluate the EDTC on two UAV tracking datasets DUT Anti-UAV~\cite{zhao2022vision} and Anti-UAV~\cite{jiang2021anti}, in comparison with several state-of-the-art trackers. 
The DUT Anti-UAV contains $20$ videos, and the Anti-UAV test set contains $91$ videos. The thermal videos of Anti-UAV dataset are used for evaluation.
The results are presented in Tab.~\ref{comparative2}.
As shown, on the DUT Anti-UAV dataset, the EDTC, whose tracking branch is trained on general tracking datasets, achieves an Acc of $0.593$. 
When trained with the additional AntiUAV600 dataset, the EDTC$^\star$ achieves an improvement to $0.606$ on Acc. 
This improvement highlights the significance of the proposed AntiUAV600 dataset for enhancing UAV perception.
Similarly, on the Anti-UAV dataset, the EDTC method outperforms all other compared trackers, achieving an Acc of $0.617$.
With additional UAV data for training, the EDTC$^\star$ obtains a higher Acc of $0.634$.
These results show the effectiveness of the proposed method in tackling the challenges of UAV perception, even in scenarios where the initial UAV state is not provided.

\textbf{Performance on the Scenarios of UAV Absence and Reappearance.} 
We select $63$ videos from the AntiUAV600 test set that feature scenarios where the UAV undergoes intervals of absence and reappearance. 
In these videos, the UAV targets are initially present in the first frame. 
The results of the proposed method, as well as existing state-of-the-art trackers, on thses videos, are presented in Tab.~\ref{results_asence}.
From the table, it is evident that the existing tracking methods struggle to handle the realistic challenge of targets reappearing after a long period of absence.
The proposed EDTC method, through the evidential collaboration of detection and tracking, demonstrates its effectiveness in handling these challenging scenarios.
Besides, compared to training the tracking branch on only general tracking datasets, the additional AntiUAV600 training set led to an improvement from $0.371$ to $0.387$ on Acc.


\subsection{Ablation Studies}

\begin{table}[!t]
\centering
\caption{Comparison of traditional tracker and UAV-specific trackers.}\label{ablation1}
\resizebox{0.65\linewidth}{!}{
\begin{tabular}{c|ccc|ccc}
\hline
\multicolumn{1}{c|}{\multirow{2}{*}{Method (YOLO+)}} & \multicolumn{3}{c|}{Simple Combination} &\multicolumn{3}{c}{Evidential Combination} \\
\cline{2-7}
 & Tracker1 & Tracker2 & Tracker3 & Tracker1 & Tracker2 & Tracker3 \\
\hline 
Acc & 0.263 & 0.278 & 0.352 & 0.439 & 0.449 & 0.486\\
\hline
\end{tabular}}
\end{table}

\textbf{Effectiveness of the AntiUAV600 Dataset.}
Firstly, experiments are conducted to demonstrate the effectiveness of the proposed AntiUAV600 dataset for UAV perception.
We construct three trackers, \emph{i.e.}, Tracker1, Tracker2 and Tracker3, trained with different datasets.
The Tracker1 is the tracking branch trained using only four general tracking datasets, while the Tracker2 and Tracker3 denote the tracker trained with additional Anti-UAV~\cite{jiang2021anti} dataset and the proposed AntiUAV600 dataset, respectively. 
To avoid duplication, we removed overlapping sequences from the Anti-UAV training set, which shares data with the AntiUAV600 test set, before conducting the training of the Tracker2.
The results are presented in Tab.~\ref{ablation1}.
The Simple Combination (SC) denotes a UAV detector first detects the initial UAV target, followed by tracking with a separate tracker.
The Evidential Combination (EC) is the proposed adaptive collaboration of the detector and tracker.
As shown, with additional training data from AntiUAV dataset and AntiUAV600 dataset, the SC and EC achieves significant performance improvements.
Specifically, comparing Tracker2 to Tracker1, we observe notable improvements on Acc, with the SC and EC increasing from $0.263$ to $0.278$ and from $0.439$ to $0.449$, respectively, thanks to the additional training provided by the Anti-UAV dataset.
Besides, the AntiUAV600 dataset significantly improves the performance of Tracker3 compared to Tracker1, with the SC increasing to $0.352$ and the EC improving to $0.486$.
Furthermore, when comparing Tracker3 to Tracker2, the AntiUAV600 dataset brings a greater improvement in both the SC and EC approaches than the Anti-UAV dataset, emphasizing its advantageous contribution to UAV perception.

\begin{wraptable}{r}{5.0cm}
\centering
\vspace{-0.6cm}
\caption{Ablation results.}\label{ablation}
\resizebox{2.0in}{!}{
\begin{tabular}{cccc}
\hline YOLO & BaseTracker & EH  & Acc \\
\hline 
\ding{52} & \ding{56} & \ding{56} & 0.392\\
\ding{52} & \ding{52} & \ding{56} & 0.352\\
\ding{52} & \ding{52} & \ding{52} & 0.486\\
\hline
\end{tabular}}
\end{wraptable}

\textbf{Effectiveness of Each Component.}
We perform ablation studies to demonstrate the effectiveness of each key component in the proposed anti-UAV method EDTC, including the global detection branch (YOLO), tracking branch and Evidential Head (EH).
The results are presented in Tab.~\ref{ablation}.
The BaseTracker is the tracking branch without the evidential head.
As shown in Tab.~\ref{ablation}, YOLO achieves $0.392$ on Acc. 
However, the simple combination of global detection and local tracking, specifically switching to local tracking after global detection, significantly decreases performance, dropping from $0.392$ to $0.352$.
This is mainly because the background of the UAV dataset is usually complex, which makes it easy for the detector to regard a distractor as a UAV object.
Once the detector makes a false positive, the tracker will track a distractor in all subsequent frames with a wrong initial state.
Therefore, a reasonable and efficient switch between the detector and tracker is crucial.
As can be seen, the integration of global detection and local tracking through the evidential head achieves significant performance improvement compared to solely relying on global detection, from $0.392$ to $0.486$ in terms of Acc.
From the above results, the proposed evidential collaboration of detection and tracking enables us to use the detector and the proposed tracker best.

\begin{wrapfigure}{r}{7.5cm}
\begin{center}
\includegraphics[trim={30mm 39mm 25mm 32mm},clip,width=0.95\linewidth]{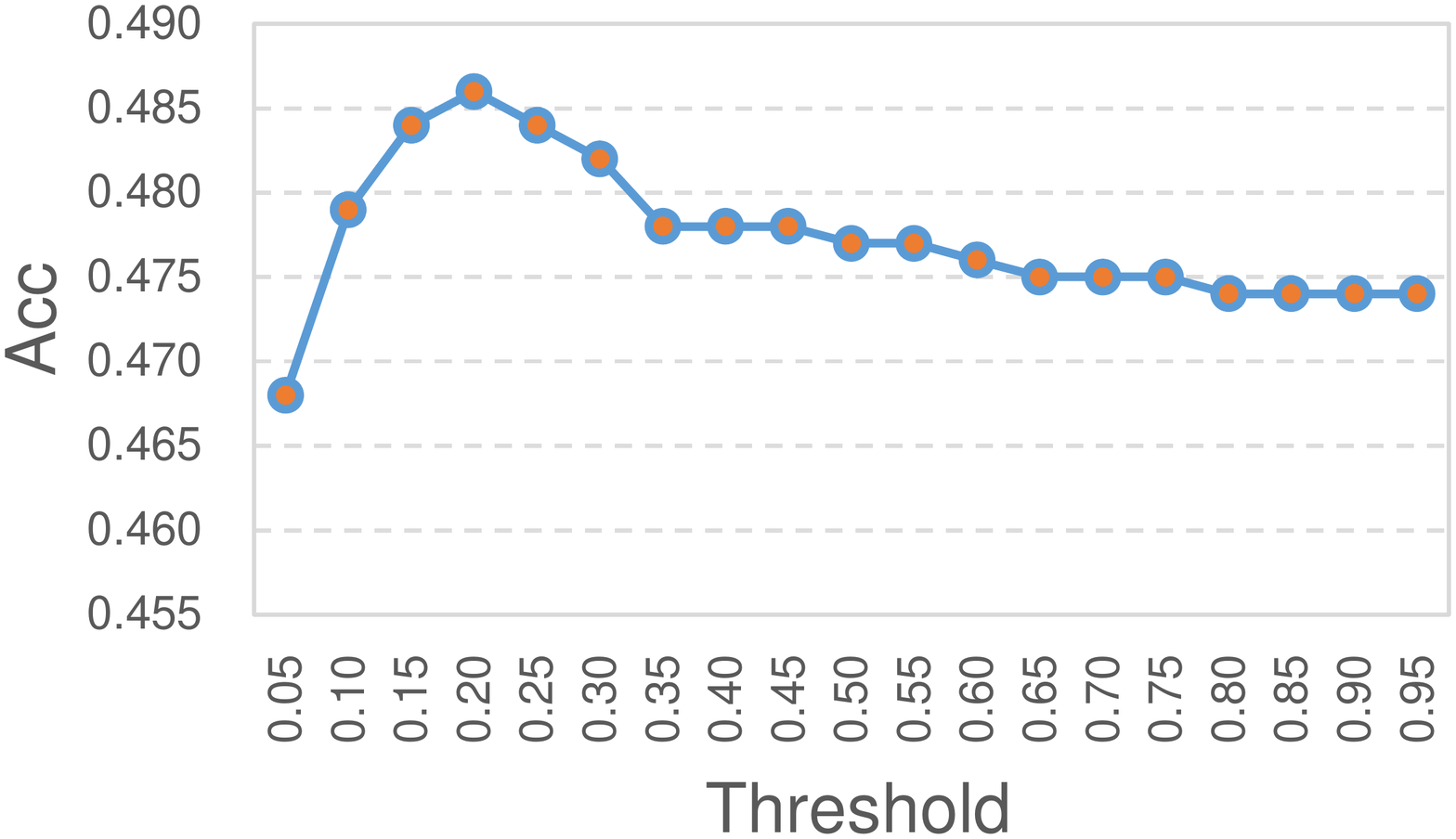}
\end{center}
\caption{Performance of the proposed evidential head with different uncertainty thresholds. 
}\label{uncertainty}
\vspace{-0.3cm}
\end{wrapfigure}

\textbf{Hyper-parameters Analysis.} The thresholds of predictive uncertainty $\theta_{eh}$ are analysed. 
The impact of the thresholds $\theta_{eh}$ is presented in the Figure~\ref{uncertainty}.
The evidential head produces a class probability and an uncertainty score.
Once the class probability indicates a background class or a target class with high uncertainty, the method will switch to global detection.
The result with a predictive target class and high uncertainty above a threshold $\theta_{eh}$ represents a possible false positive prediction. 
The best performance of $0.486$ is achieved when $\theta_{eh}$ is set as $0.2$ in the Figure~\ref{uncertainty}, as true positive tracking results typically exhibit low uncertainty.
With the increase of the $\theta_{eh}$, the performance drops, highlighting the accuracy of the predicted uncertainty of the evidential head.
Additionally, the figure illustrates that even with a low threshold of $0.05$, \emph{i.e.}, only utilizing the classification results from the evidential head for decision-making, there is a significant improvement in performance, from $0.352$ to $0.468$.

\section{Conclusion}
\label{conlusion}

This study presents a more practical and comprehensive problem formulation for UAV perception.
The newly formulated task entails the anti-UAV systems to continuously detect the presence of UAVs and accurately localize the UAV targets, relying only on the real-time captured videos.
To support the new task, we propose a novel benchmark including a new large-scale dataset AntiUAV600 and a novel performance measure are pospoed.
In contrast to existing UAV datasets that contain limited annotated frames with provided initial state of the UAV, the AntiUAV600 offers more realistic scenarios. 
Besides, the proposed new performance measure takes into account both the localization accuracy and the ability to predict the UAV absence.
To serve as a baseline for future research and inspire more solutions, we propose an approach that combines detection and tracking algorithms, where the global detection stage and local tracking stage are dynamically switched by an evidential learning module.
Furthermore, the comprehensive experiments and analyses reported have demonstrated the effectiveness and advantage of the proposed solution and benchmark.

\bibliographystyle{plain}
\bibliography{neurips_data_2023}

\begin{thebibliography}{10}

\bibitem{amini2020deep}
Alexander Amini, Wilko Schwarting, Ava Soleimany, and Daniela Rus.
\newblock Deep evidential regression.
\newblock In {\em NeurIPS}, pages 14927--14937, 2020.

\bibitem{bao2021evidential}
Wentao Bao, Qi~Yu, and Yu~Kong.
\newblock Evidential deep learning for open set action recognition.
\newblock In {\em ICCV}, pages 13349--13358, 2021.

\bibitem{bhat2019learning}
Goutam Bhat, Martin Danelljan, Luc~Van Gool, and Radu Timofte.
\newblock Learning discriminative model prediction for tracking.
\newblock In {\em ICCV}, pages 6182--6191, 2019.

\bibitem{chen2017deep}
Yueru Chen, Pranav Aggarwal, Jongmoo Choi, and C-C~Jay Kuo.
\newblock A deep learning approach to drone monitoring.
\newblock In {\em APSIPA ASC}, pages 686--691, 2017.

\bibitem{coluccia2021drone}
Angelo Coluccia, Alessio Fascista, Arne Schumann, Lars Sommer, Anastasios
  Dimou, Dimitrios Zarpalas, Fatih~Cagatay Akyon, Ogulcan Eryuksel, Kamil~Anil
  Ozfuttu, Sinan~Onur Altinuc, et~al.
\newblock Drone-vs-bird detection challenge at ieee avss2021.
\newblock In {\em AVSS}, pages 1--8, 2021.

\bibitem{cui2022mixformer}
Yutao Cui, Cheng Jiang, Limin Wang, and Gangshan Wu.
\newblock Mixformer: End-to-end tracking with iterative mixed attention.
\newblock In {\em CVPR}, pages 13608--13618, 2022.

\bibitem{dadboud2021single}
Fardad Dadboud, Vaibhav Patel, Varun Mehta, Miodrag Bolic, and Iraj Mantegh.
\newblock Single-stage uav detection and classification with yolov5: Mosaic
  data augmentation and panet.
\newblock In {\em AVSS}, pages 1--8, 2021.

\bibitem{del2021unmanned}
Jaime del Cerro, Christyan Cruz~Ulloa, Antonio Barrientos, and Jorge
  de~Le{\'o}n~Rivas.
\newblock Unmanned aerial vehicles in agriculture: A survey.
\newblock {\em Agronomy}, 11(2):203, 2021.

\bibitem{deng2009imagenet}
Jia Deng, Wei Dong, Richard Socher, Li-Jia Li, Kai Li, and Li~Fei-Fei.
\newblock Imagenet: A large-scale hierarchical image database.
\newblock In {\em CVPR}, pages 248--255, 2009.

\bibitem{fan2019lasot}
Heng Fan, Liting Lin, Fan Yang, Peng Chu, Ge~Deng, Sijia Yu, Hexin Bai, Yong
  Xu, Chunyuan Liao, and Haibin Ling.
\newblock Lasot: A high-quality benchmark for large-scale single object
  tracking.
\newblock In {\em CVPR}, pages 5374--5383, 2019.

\bibitem{guo2020siamcar}
Dongyan Guo, Jun Wang, Ying Cui, Zhenhua Wang, and Shengyong Chen.
\newblock Siamcar: Siamese fully convolutional classification and regression
  for visual tracking.
\newblock In {\em CVPR}, pages 6269--6277, 2020.

\bibitem{huang2021siamsta}
Bo~Huang, Junjie Chen, Tingfa Xu, Ying Wang, Shenwang Jiang, Yuncheng Wang, Lei
  Wang, and Jianan Li.
\newblock Siamsta: Spatio-temporal attention based siamese tracker for tracking
  uavs.
\newblock In {\em ICCVW}, pages 1204--1212, 2021.

\bibitem{huang2019got}
Lianghua Huang, Xin Zhao, and Kaiqi Huang.
\newblock Got-10k: A large high-diversity benchmark for generic object tracking
  in the wild.
\newblock {\em IEEE T-PAMI}, 43(5):1562--1577, 2019.

\bibitem{isaac2021unmanned}
Brian~KS Isaac-Medina, Matt Poyser, Daniel Organisciak, Chris~G Willcocks,
  Toby~P Breckon, and Hubert~PH Shum.
\newblock Unmanned aerial vehicle visual detection and tracking using deep
  neural networks: A performance benchmark.
\newblock In {\em ICCVW}, pages 1223--1232, 2021.

\bibitem{jiang2021anti}
Nan Jiang, Kuiran Wang, Xiaoke Peng, Xuehui Yu, Qiang Wang, Junliang Xing,
  Guorong Li, Qixiang Ye, Jianbin Jiao, Zhenjun Han, et~al.
\newblock Anti-uav: a large-scale benchmark for vision-based uav tracking.
\newblock {\em IEEE T-MM}, 25:486--500, 2021.

\bibitem{li2023global}
Yifan Li, Dian Yuan, Meng Sun, Hongyu Wang, Xiaotao Liu, and Jing Liu.
\newblock A global-local tracking framework driven by both motion and
  appearance for infrared anti-uav.
\newblock In {\em CVPRW}, pages 3025--3034, 2023.

\bibitem{lin2014microsoft}
Tsung-Yi Lin, Michael Maire, Serge Belongie, James Hays, Pietro Perona, Deva
  Ramanan, Piotr Doll{\'a}r, and C~Lawrence Zitnick.
\newblock Microsoft coco: Common objects in context.
\newblock In {\em ECCV}, pages 740--755, 2014.

\bibitem{mayer2022transforming}
Christoph Mayer, Martin Danelljan, Goutam Bhat, Matthieu Paul, Danda~Pani
  Paudel, Fisher Yu, and Luc Van~Gool.
\newblock Transforming model prediction for tracking.
\newblock In {\em CVPR}, pages 8731--8740, 2022.

\bibitem{mayer2021learning}
Christoph Mayer, Martin Danelljan, Danda~Pani Paudel, and Luc Van~Gool.
\newblock Learning target candidate association to keep track of what not to
  track.
\newblock In {\em CVPR}, pages 13444--13454, 2021.

\bibitem{muller2018trackingnet}
Matthias Muller, Adel Bibi, Silvio Giancola, Salman Alsubaihi, and Bernard
  Ghanem.
\newblock Trackingnet: A large-scale dataset and benchmark for object tracking
  in the wild.
\newblock In {\em ECCV}, pages 300--317, 2018.

\bibitem{partheepan2023autonomous}
Shouthiri Partheepan, Farzad Sanati, and Jahan Hassan.
\newblock Autonomous unmanned aerial vehicles in bushfire management:
  Challenges and opportunities.
\newblock {\em Drones}, 7(1):47, 2023.

\bibitem{redmon2016you}
Joseph Redmon, Santosh Divvala, Ross Girshick, and Ali Farhadi.
\newblock You only look once: Unified, real-time object detection.
\newblock In {\em CVPR}, pages 779--788, 2016.

\bibitem{rodriguez2020adaptive}
Alejandro Rodriguez-Ramos, Javier Rodriguez-Vazquez, Carlos Sampedro, and
  Pascual Campoy.
\newblock Adaptive inattentional framework for video object detection with
  reward-conditional training.
\newblock {\em IEEE Access}, 8:124451--124466, 2020.

\bibitem{sensoy2020uncertainty}
Murat Sensoy, Lance Kaplan, Federico Cerutti, and Maryam Saleki.
\newblock Uncertainty-aware deep classifiers using generative models.
\newblock In {\em AAAI}, volume~34, pages 5620--5627, 2020.

\bibitem{sensoy2018evidential}
Murat Sensoy, Lance Kaplan, and Melih Kandemir.
\newblock Evidential deep learning to quantify classification uncertainty.
\newblock In {\em NeurIPS}, pages 01--11, 2018.

\bibitem{shi2020multifaceted}
Weishi Shi, Xujiang Zhao, Feng Chen, and Qi~Yu.
\newblock Multifaceted uncertainty estimation for label-efficient deep
  learning.
\newblock In {\em NeurIPS}, pages 17247--17257, 2020.

\bibitem{vskrinjar2019application}
Jasmina~Pa{\v{s}}agi{\'c} {\v{S}}krinjar, Pero {\v{S}}korput, and Martina
  Furdi{\'c}.
\newblock Application of unmanned aerial vehicles in logistic processes.
\newblock In {\em NTDA IV}, pages 359--366, 2019.

\bibitem{svanstrom2021real}
Fredrik Svanstr{\"o}m, Cristofer Englund, and Fernando Alonso-Fernandez.
\newblock Real-time drone detection and tracking with visible, thermal and
  acoustic sensors.
\newblock In {\em ICPR}, pages 7265--7272, 2021.

\bibitem{wu2021cvt}
Haiping Wu, Bin Xiao, Noel Codella, Mengchen Liu, Xiyang Dai, Lu~Yuan, and Lei
  Zhang.
\newblock Cvt: Introducing convolutions to vision transformers.
\newblock In {\em ICCV}, pages 22--31, 2021.

\bibitem{xie2023stftrack}
Xueli Xie, Jianxiang Xi, Xiaogang Yang, Ruitao Lu, and Wenxin Xia.
\newblock Stftrack: Spatio-temporal-focused siamese network for infrared uav
  tracking.
\newblock {\em Drones}, 7(5):296, 2023.

\bibitem{yan2021learning}
Bin Yan, Houwen Peng, Jianlong Fu, Dong Wang, and Huchuan Lu.
\newblock Learning spatio-temporal transformer for visual tracking.
\newblock In {\em ICCV}, pages 10448--10457, 2021.

\bibitem{ye2022joint}
Botao Ye, Hong Chang, Bingpeng Ma, Shiguang Shan, and Xilin Chen.
\newblock Joint feature learning and relation modeling for tracking: A
  one-stream framework.
\newblock In {\em ECCV}, pages 341--357. Springer, 2022.

\bibitem{yu2023unified}
Qianjin Yu, Yinchao Ma, Jianfeng He, Dawei Yang, and Tianzhu Zhang.
\newblock A unified transformer based tracker for anti-uav tracking.
\newblock In {\em CVPRW}, pages 3035--3045, 2023.

\bibitem{zhao20233rd}
Jian Zhao, Jianan Li, Lei Jin, Jiaming Chu, Zhihao Zhang, Jun Wang, Jiangqiang
  Xia, Kai Wang, Yang Liu, Sadaf Gulshad, et~al.
\newblock The 3rd anti-uav workshop \& challenge: Methods and results.
\newblock {\em arXiv preprint arXiv:2305.07290}, 2023.

\bibitem{zhao20212nd}
Jian Zhao, Gang Wang, Jianan Li, Lei Jin, Nana Fan, Min Wang, Xiaojuan Wang,
  Ting Yong, Yafeng Deng, Yandong Guo, et~al.
\newblock The 2nd anti-uav workshop \& challenge: methods and results.
\newblock {\em arXiv preprint arXiv:2108.09909}, pages 1--8, 2021.

\bibitem{zhao2022vision}
Jie Zhao, Jingshu Zhang, Dongdong Li, and Dong Wang.
\newblock Vision-based anti-uav detection and tracking.
\newblock {\em IEEE T-ITS}, 23(12):25323--25334, 2022.

\end{thebibliography}

\clearpage
\begin{@twocolumnfalse}
	\section*{\centering{\textbf{Supplementary Material for \\ \emph{Evidential Detection and Tracking Collaboration: New Problem, \\Benchmark and Algorithm for Robust Anti-UAV System}}\\[35pt]}}
\end{@twocolumnfalse}

\begin{appendices}
In this supplementary material, we provide additional details of the proposed AntiUAV600 dataset, experimental implementation details, and qualitative results.

\begin{figure}[h]
\begin{center}
\includegraphics[trim={25mm 34mm 25mm 0mm},clip,width=1\linewidth]{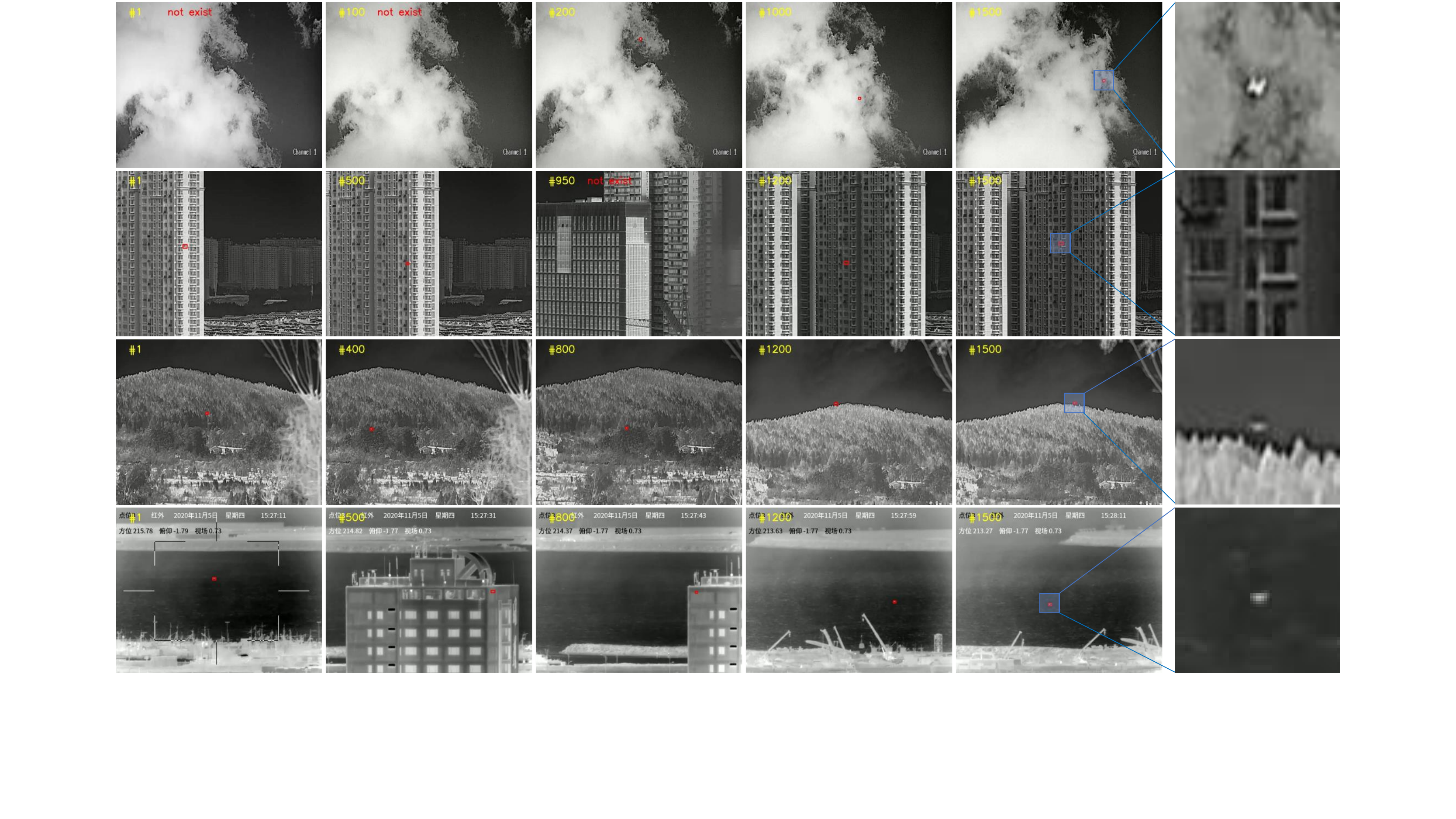}
\end{center}
\caption{Sequences samples from the proposed AntiUAV600 dataset. Annotations are shown in red color.}\label{samples}
\end{figure}

\section{Dataset Details}
\begin{wrapfigure}{r}{6.2cm}
\begin{center}
\vspace{-0.6cm}
\includegraphics[trim={18mm 100mm 18mm 112mm},clip,width=0.95\linewidth]{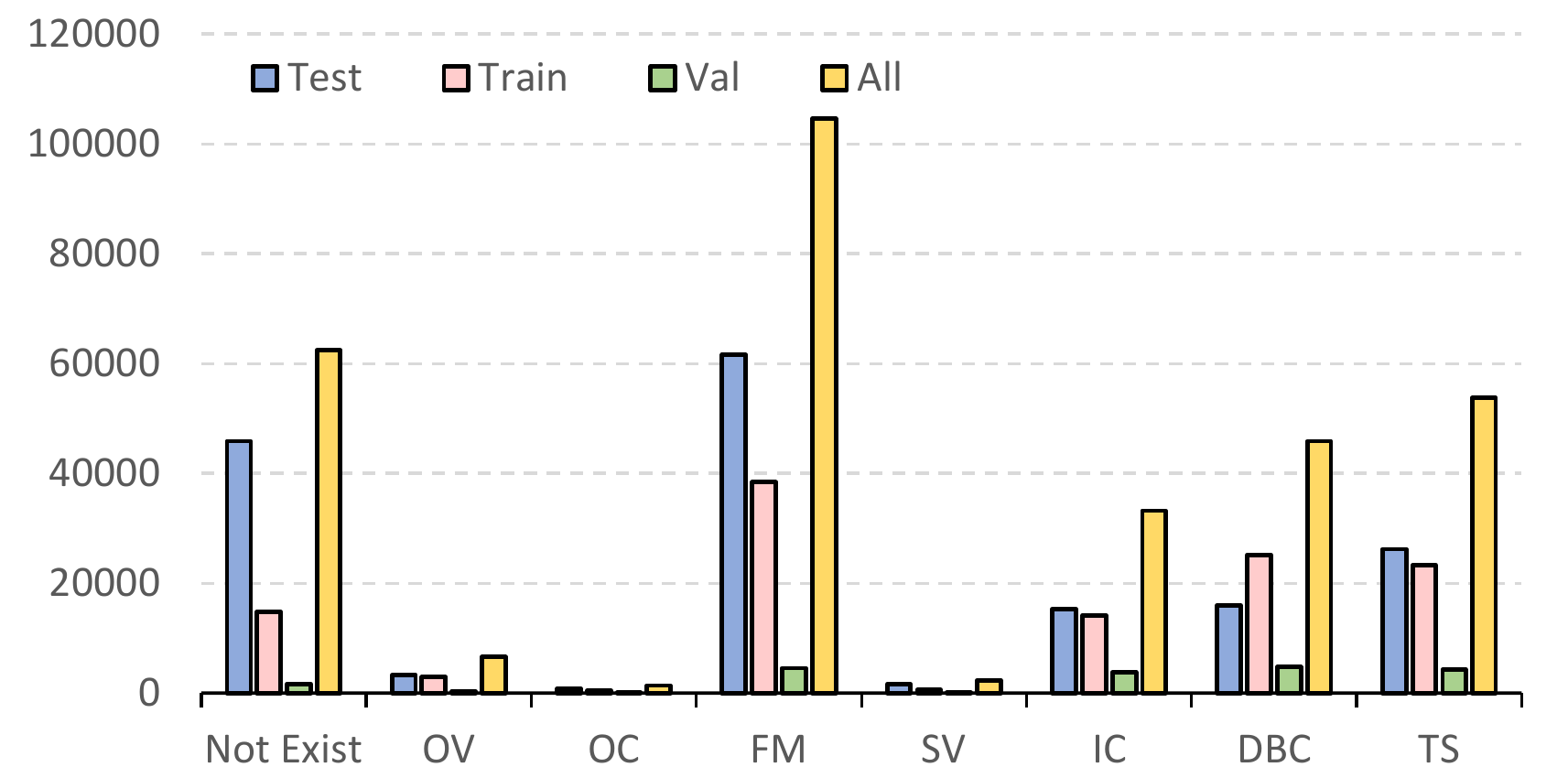}
\end{center}
\caption{The distribution of frames with different attributes in the AntiUAV600 dataset.}\label{distribution}
\end{wrapfigure}
\textbf{Sequence samples.}
In Figure~\ref{samples}, several video sequences of the AntiUAV600 dataset are presented.
The sequences in the AntiUAV600 dataset encompass diverse and challenging background scenes, such as cloudy sky, urban environments, mountains, rivers, and forests. 
Besides, these videos present a wide range of challenges, including UAV absence and reappearance, fast motion, dynamic backgrounds, infrared crossover, and small scales. 
The presence of these interfering factors in the dataset emphasizes its practical and challenging nature for anti-UAV tasks.
In addition, as shown in the first row of sequences, in a practical anti-UAV scenario, the UAV target does not always appear in the first frame.
This kind of sequence cannot be processed solely using standard tracking techniques, which inspired us to propose an anti-UAV algorithm that involves the collaboration of detection and tracking.

\textbf{Challenging Attributes.}
All frames are annotated with seven challenging attributes, including Out-of-View (OV), Occlusion (OC), Fast Motion (FM), Scale Variation (SV), Infrared Crossover (IC), Dynamic Background Clusters (DBC), and Target Scale (TS). The definitions of these attributes are listed below:
\begin{itemize}[itemindent=0.3cm, itemsep=1.0mm]
\item Out-of-View (OV): the target moves out of the current view.
\item Occlusion (OC): the target is partially or heavily occluded.
\item Fast Motion (FM): the target moves quickly.
\item Scale Variation (SV): the scale of the bounding boxes over the frames varies significantly.
\item Infrared Crossover (IC): the target has a similar temperature to other objects or background surroundings.
\item Dynamic Background Clusters (DBC): there are dynamic changes (e.g., waves, leaves, birds) in the background around the target.
\item Target Scale (TS): the target is with a tiny, small, medium, or large scale.
\end{itemize}
Besides, in Figure~\ref{distribution}, the distribution of the frames with seven challenging attributes and target absence is provided.
The figure shows that frequent absence, fast motion, thermal crossover, background clutter, and tiny size are the main challenges for UAV perception in the AntiUAV600 dataset.

\section{Experiments}
\subsection{Implementation Details}
\textbf{Architectures.}
The proposed anti-UAV framework is composed of a global detection branch and a local tracking branch.
The global detection branch utilizes YOLOv5s~\cite{redmon2016you}, a lightweight detector.
As to the local tracking branch, given a template with the size of ${H_t}\times{W_t}\times3$ and a search region with the size of $H_s\times{W_s}\times3$, the number of tokens for the template and search region are $({\frac{H_t}{4}}\times\frac{W_t}{4}+\frac{H_s}{4}\times\frac{W_s}{4})\times{C}$ in the first stage.
Then, the tokens are fed into the relevance decoupling module for feature extraction and matching.
After the 3-stage patch embedding and relevance decoupling module, the dimension of the features is ${(\frac{H_t}{16}}\times\frac{W_t}{16}+\frac{H_s}{16}\times\frac{W_s}{16})\times{6C}$.
Finally, the search region features with the size of $\frac{H_s}{16}\times\frac{W_s}{16}\times{6C}$ are fed into the corner prediction head~\cite{yan2021learning} to estimate the target position and scale.
The corner prediction head is composed of several fully-convolutional layers to predict the top-left and bottom-right corners of the target bounding box.

\textbf{Training}
The detection branch and the tracking branch are trained separately.
The YOLOv5s model is trained on the AntiUAV600 training set for 20 epochs using three RTX2080Ti GPUs, and the tracking model is trained in two stages using 4 RTX3090Ti GPUs on our AntiUAV600 dataset and four general tracking datasets, including LaSOT~\cite{fan2019lasot}, TrackingNet~\cite{muller2018trackingnet}, GOT10K~\cite{huang2019got}, and COCO~\cite{lin2014microsoft}.
In the first stage, we train the backbone and prediction head of the tracking model for 500 epochs, with the learning rate initialized as 1e-4 and decreased to 1e-5 at the 400-th epoch.
The backbone is initialized with the CVT-21~\cite{wu2021cvt} pre-trained on the ImageNet dataset~\cite{deng2009imagenet}.
The scales of template and search region are set to $128\times128$ and $320\times320$ respectively, and $C$ is set as $64$.
In the second stage, the evidential head is trained for 40 epochs. 
We set $\lambda_{iou}=2$ and $\lambda_{L_1}=5$ for training.

\textbf{Inference.}
The proposed EDTC method performs continuous detection and tracking using only the entire image as input.
Specifically, for global detection, the entire image is input to the detector to detect the UAV target.
The detector prediction with the highest score is selected as the UAV target.
For the tracking branch, the detected UAV target template and the current search region are the input of the tracking branch.
In particular, when testing the proposed method on DUT Anti-UAV~\cite{zhao2022vision} (Table. 3), the detector of EDTC is trained on the DUT Anti-UAV detection subset.
Similarly, for the evaluation on Anti-UAV~\cite{jiang2021anti} (Table. 3), due to the presence of data overlap between the proposed AntiUAV600 dataset and the Anti-UAV dataset, the detector is trained on Anti-UAV training set.
The tracking branch is trained on four general tracking datasets and the Anti-UAV training set.

\subsection{Qualitative Results.}
\begin{figure}[t]
\begin{center}
\includegraphics[trim={0mm 80mm 80mm 00mm},clip,width=0.95\linewidth]{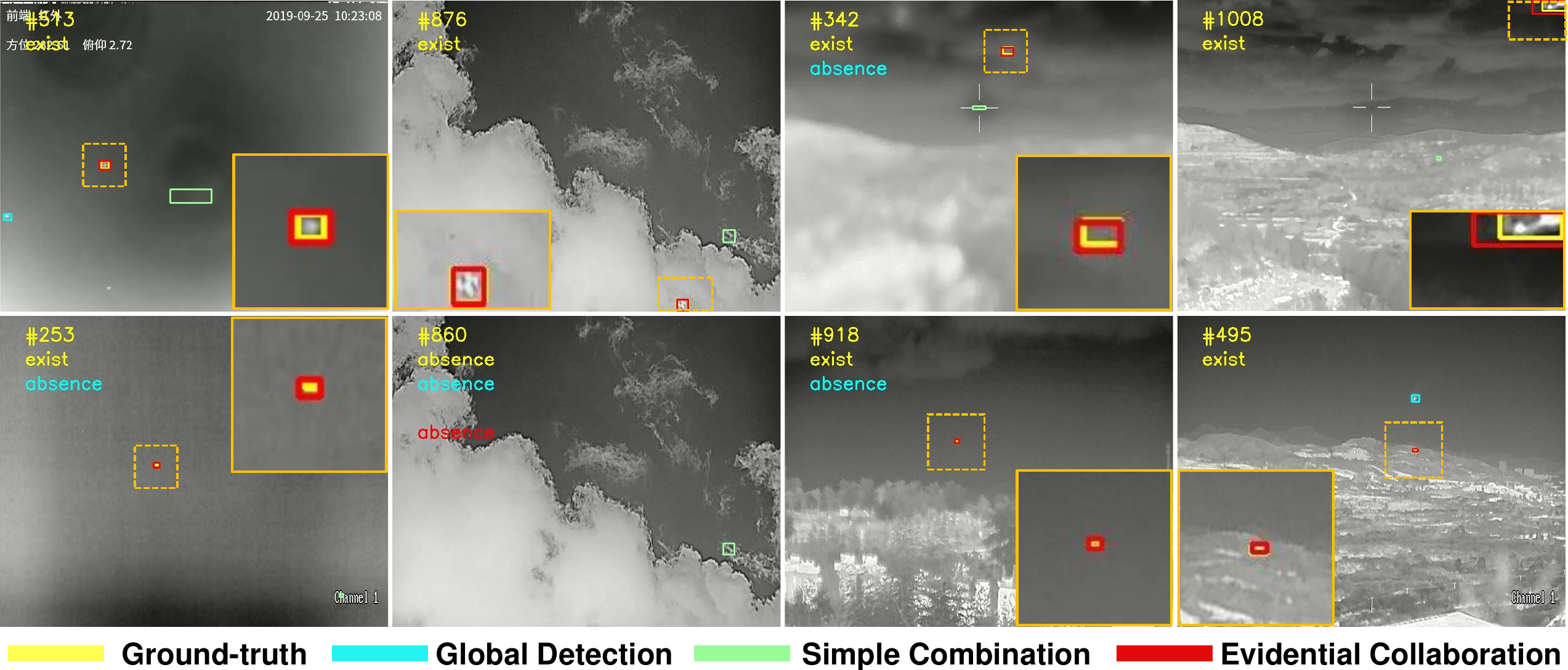}
\end{center}
\caption{Illustration of qualitative experimental results on several challenging sequences. The results of the proposed evidential collaboration of detection and tracking, global detection, a simple combination of detection and tracking, and the ground-truth bounding boxes are provided. The UAV target area is zoomed in to provide a clearer view.}\label{qualitative}
\end{figure}

To intuitively demonstrate the advantages of our formulation, in Figure.~\ref{qualitative}, we provide qualitative results of our solution, global detection YOLOv5s, and a simple combination of detector and tracker.
The proposed method achieves more accurate and steady UAV localization.
As shown, although these samples are challenging with target absence and reappearance, tiny scale, infrared crossover, and clutter background, the proposed evidential detection and tracking collaboration enables us to effectively take advantage of both detection and tracking, resulting in improved UAV perception performance.

\end{appendices}

\end{document}